\documentclass[journal]{IEEEtran}

\ifCLASSINFOpdf

\else

\fi

\usepackage{amssymb}
\usepackage{amsmath}
\usepackage{graphicx}
\usepackage{booktabs}
\usepackage{xspace}
\usepackage{subfigure}
\usepackage{multirow}
\usepackage{algorithm}
\usepackage{algorithmic}
\usepackage{cite}
\usepackage{amssymb}
\usepackage{amsmath}
\usepackage[numbers,sort&compress]{natbib}
\usepackage{caption}
\graphicspath{{figures/}}
\usepackage[colorlinks,citecolor=black,urlcolor=black,linkcolor=black]{hyperref}
\usepackage[OT1]{fontenc} 

\usepackage{fancyhdr}

\begin{document}

\title{Infrared Small-Dim Target Detection with Transformer under Complex Backgrounds}

\author{Fangcen Liu,
        Chenqiang Gao*,
        Fang Chen,
        Deyu Meng, \IEEEmembership{Member, IEEE},
        Wangmeng Zuo, \IEEEmembership{Senior Member, IEEE},
        Xinbo Gao, \IEEEmembership{Senior Member, IEEE}
\thanks{*Corresponding author: Chenqiang Gao.}

\thanks{Fangcen Liu,  Chenqiang Gao, Xinbo Gao are with the School of Communication and Information Engineering, Chongqing University of Posts and Telecommunications, and also with Chongqing Key Laboratory of Signal and Information Processing, Chongqing University of Posts and Telecommunications, Chongqing 400065, China (e-mail: liufc67@gmail.com, gaocq@cqupt.edu.cn, gaoxb@cqupt.edu.cn).}

\thanks{Fang Chen is with the Viterbi School of Engineering, University of Southern California, California 90089, USA (e-mail: fchen905@usc.edu)}
 
\thanks{Deyu Meng is with School of Mathematics and Statistics, Xi’an Jiaotong University, Xi’an, Shanxi, 710049, China, and also with Macau Institute of Systems Engineering, Macau University of Science and Technology, Taipa, 999078, Macau (e-mail: dymeng@mail.xjtu.edu.cn).}
 
\thanks{Wangmeng Zuo is with School of Computer Science and Technology, Harbin Institute of Techonlogy, 47822 Harbin, Heilongjiang, China (e-mail: wmzuo@hit.edu.cn).}}

\maketitle

\begin{abstract}
The infrared small-dim target detection is one of the key techniques in the infrared search and tracking system.
Since the local regions similar to infrared small-dim targets spread over the whole background, exploring the interaction information amongst image features in large-range dependencies to mine the difference between the target and background is crucial for robust detection.
However, existing deep learning-based methods are limited by the locality of convolutional neural networks, which impairs the ability to capture large-range dependencies.
Additionally, the small-dim appearance of the infrared target makes the detection model highly possible to miss detection.
To this end, we propose a robust and general infrared small-dim target detection method with the transformer.
We adopt the self-attention mechanism of the transformer to learn the interaction information of image features in a larger range.
Moreover, we design a feature enhancement module to learn discriminative features of small-dim targets to avoid miss detection.
After that, to avoid the loss of the target information, we adopt a decoder with the U-Net-like skip connection operation to contain more information of small-dim targets.
Finally, we get the detection result by a segmentation head.
Extensive experiments on two public datasets show the obvious superiority of the proposed method over state-of-the-art methods and the proposed method has stronger cross-scene generalization and anti-noise performance.

\end{abstract}


\IEEEpeerreviewmaketitle

\section{Introduction}
\label{sec:introduction}
\begin{figure}
    \centering
    \includegraphics[width=0.48\textwidth]{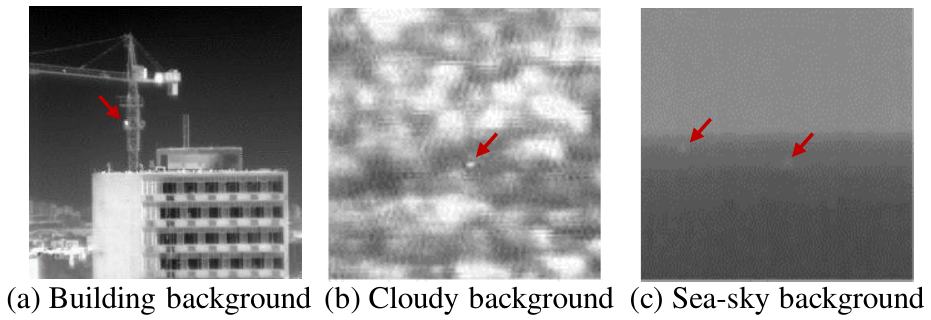}
    \caption{Illustration of challenges of the infrared small-dim target detection task under complex backgrounds.
    These targets in different scenes appear pretty small-dim and sparse, which makes the detection model easy to miss detection.
    From (a) to (c), we can also obviously observe that the local regions similar to infrared small-dim targets spread over the whole background, which easily leads to high false alarms.    
    }
    \label{fig:diff}
\end{figure}

The infrared small-dim target detection is one of the key techniques in the infrared search and tracking (IRST) system because the infrared imaging can capture targets from a long distance and has a strong anti-interference ability \cite{zhu2020balanced, zhang2021infrared, lu2020robust}. 
However, this task encounters kinds of challenges \cite{wang2019miss}, as illustrated in Fig. \ref{fig:diff}.
Infrared targets in infrared images are small and dim, while backgrounds are usually complex. 
As a result, the small-dim target is easily submerged in the complex background, with a low Signal-to-Clutter Ratio (SCR). 
In addition, the number of target pixels is much fewer than background pixels, which leads to that the target and background pixels in an image are of extreme imbalance. 

To address above challenges, model-driven infrared small-dim target detection methods routinely design the detection model for small-dim targets by deeply mining the prior knowledge of imaging characteristics of small-dim targets, backgrounds, or both of them \cite{deng2016infrared, han2018infrared, gao2013infrared}. 
However, these approaches heavily rely on prior knowledge, which makes their generalization ability limited.
Additionally, model-driven approaches can not be fast and easily applied to a new application scene whose imaging characteristic does not well match the assumption of the model.
In contrast, data-driven models are more feasible and can easily adapt to a new application scene through sample learning.   
In recent years, deep learning methods are adopted to detect infrared small-dim targets and have shown stronger generalization ability and feasibility \cite{wang2019miss, dai2021attentional}.

However, among existing deep learning methods, feature learning mainly relies on convolutional neural networks (CNNs).
The locality of CNNs impairs the ability to capture large-range dependencies \cite{DBLP:journals/corr/abs-2103-10697}, which easily results in high false alarms. As can be observed from Fig. \ref{fig:diff}, the local regions similar to infrared small-dim targets spread over the whole background.
Thus, it is very important to learn the difference between the target and the background in a large range.
																													   
Currently, the transformer structure, from the Natural Language Processing (NLP) field \cite{ye2021contrastive, vaswani2017attention}, has demonstrated its powerful ability in non-local feature learning in various computer vision tasks \cite{nguyen2021modular, dosovitskiy2020image}. 
Different from CNN architectures, the transformer architecture contains the self-attention mechanism and the feed-forward network.
The self-attention mechanism enables the transformer to have the ability to capture large-range dependencies of all embedded tokens. 

In this paper, we adopt the transformer to learn the interaction information amongst all embedded tokens of an image. 
Firstly, we embed an image into a sequence of tokens by the Resnet-50 \cite{he2016deep}.
After that, the self-attention mechanism is adopted to model complex dependencies among different embedded tokens, so that the difference between the small-dim target and background can be well mined.

Furthermore, due to the small size and the dim appearance of the target, as can be seen from Fig. \ref{fig:diff}, if we can not capture discriminative information of small-dim targets, it highly possibly lend to miss detection.
To this end, we design a feature enhancement module as the feed-forward network to acquire more discriminative features of small-dim targets.
Moreover, since small-dim target features are easily lost in the network, we adopt a U-Net-like \cite{ronneberger2015u} upsampling structure to get more information of small-dim targets.

We summarize the main contributions of the paper as follows:
\begin{itemize}
     \item We propose a novel small-dim target detection method.
     It adopts the self-attention mechanism of the transformer to learn the interaction information amongst all embedded tokens so that the network can learn the difference between the small-dim target and background in a larger range.
     To our best knowledge, this is the first work to explore the transformer to detect the infrared small-dim target.
     \item The designed feature enhancement module can help learn more discriminative features of small-dim targets.
     \item We evaluate the proposed method on two public datasets and extensive experimental results show that the proposed method is effective and significantly outperforms state-of-the-art methods.
 \end{itemize}
 
The remainder of this paper is organized as follows:
 In Section \ref{sec:related work}, related works are briefly reviewed.
 In Section \ref{sec:method}, we present the proposed method in detail.
 In Section \ref{sec:experiment}, the experimental results are given and discussed.
 Conclusions are drawn in Section \ref{sec:conclude}.

\section{Related Work} 
\label{sec:related work}
\subsection{Small target detection}
\subsubsection{Infrared small-dim target detection}
In the early stages, the model-driven infrared small-dim target detection methods design filters to enhance the target or suppress the background \cite{han2020infrared, deng2021infrared}.
Zeng et al. \cite{zeng2006design} and Deshpande et al. \cite{deshpande1999max} proposed the Top-Hat method and max-mean/max-median method to directly enhance targets by filtering them out from original images, respectively.
Then, Deng et al. \cite{deng2021infrared} proposed an adaptive M-estimator ring top-hat transformation method to detection the small-dim target. 
Aghaziyarati et al. \cite{aghaziyarati2019small} and Moradi et al.\cite{moradi2020fast} detected small-dim targets by suppressing the estimated background as much as possible.
However, detection performances of these methods are limited by designed filters.
Inspired by the human visual system, some efforts are based on the different local contrast which focus on the saliency of the target to distinguish the target from the background and improve the performance of small-dim infrared target detection \cite{han2019local, deng2016small, gao2008generalised}.
Deng et al. \cite{deng2016small} and Gao et al. \cite{gao2008generalised} focused on the saliency of the target to distinguish the target from the background.
Cui et al. \cite{cui2016infrared} proposed an infrared small-dim target detection algorithm with two layers which can balance those detection capabilities.
The first layer was designed to select significant local information. 
Then the second layer leveraged a classifier to separate targets from background clutters.
Based on the observation that small-dim infrared targets often present sparse features, while the background has the non-local correlation property \cite{pang2021facet, sun2020infrared}. 
Gao et al. \cite{gao2013infrared} firstly proposed the patch-image in infrared small-dim target detection and the low-rank-based infrared patch-image (IPI) model. 
Dai et al. \cite{dai2016infrared} proposed a column weighted IPI model (WIPI), and then proposed a reweighed infrared patch-tensor model (RIPT) \cite{dai2017reweighted}. 
However, the model-driven approaches heavily rely on prior knowledge, which makes the generalization ability of these models limited.

Recently, the generalization ability of the data-driven infrared small-dim target detection is well promoted by deep learning methods \cite{zhao2020novel, ju2021istdet, ryu2019heterogeneous, gao2019dim, du2021cnn, tong2021eaau, shi2020infrared, hou2021ristdnet}.
Fan et al. \cite{fan2018dim} designed a convolutional neural network architecture to improve the contrast between the small-dim target and the background.
Zhao et al. \cite{zhao2019tbc} proposed a TBC-Net which included a target extraction module and semantic constraint module.
The target extraction module aimed to predict the potential target, while the semantic constraint module aimed to constrain the number of small-dim targets.
Wang et al. \cite{wang2019miss} used adversarial generation networks (GANs) to balance Miss Detection (MD) and False Alarm (FA).
Shi et al. \cite{shi2019infrared} and Zhao et al. \cite{zhao2020novel} regarded small-dim targets as noises, and they treated the small-dim targets detection task as a denoising task.
Hou et al. \cite{RISTDnet} combined handcrafted feature methods and convolutional neural networks and proposed a robust RISTDnet framework.
Dai et al. \cite{dai2021attentional, dai21acm} combined the global and local information of the infrared image and proposed end-to-end detection methods named ALC and ACM to solve the problem of the lack of fixed features of infrared small-dim targets.

Compared with above data-driven methods, the proposed method in this paper overcomes the shortcoming of the limitation of the CNN to learn the interaction information amongst all embedded tokens, which can mine differences between targets and backgrounds in a larger range.

\begin{figure*}
    \centering
    \includegraphics[width=1\textwidth]{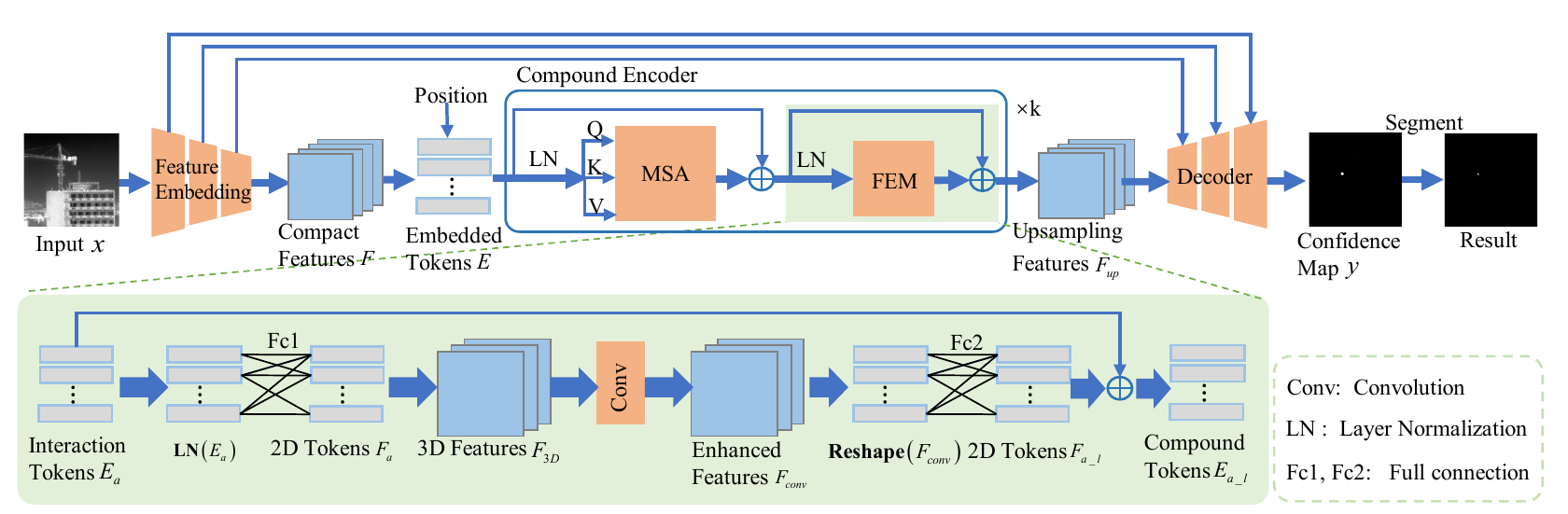}
    \caption{The proposed infrared small-dim target detection framework in this paper.
    It contains three parts: a feature embedding module, a compound encoder with $k$ encoder layers, and a decoder.
    The feature embedding module is proposed to obtain the compact feature representation.
    In the compound encoder, each encoder layer mainly has two parts: the multi-head self-attention (MSA) mechanism and the feature enhancement module (FEM).
    The multi-head self-attention is adopted to learn the interaction information amongst all embedded tokens.
    The designed feature enhancement module can learn more discriminative features of small-dim targets.
    The decoder is adopted to obtain the confidence map of the small-dim target.
    Finally, we obtain the detection result through the adaptive threshold segmentation.
    }
    \label{fig:pipline}
 \end{figure*}

\subsubsection{Small target detection in RGB images}
Different from infrared small-dim target methods, small target detection methods for RGB images payed more attention to solving the problem of small target size \cite{Uzair2021, Zhou2021}, and usually adopted data augmentation \cite{zoph2020learning}, multi-scale learning \cite{singh2018analysis} and context information learning \cite{torralba2001statistical} strategies to improve the robustness and generalization of detection.
However, using the above methods directly to detect infrared small-dim targets would make the performance drop sharply, as verified in \cite{wang2019miss}.
Compared with targets in RGB images, infrared small-dim targets have high similarity to the background with low SCR, so it is more difficult to distinguish small-dim targets from backgrounds.
Moreover, the max-pooling layer adopted in these methods may suppress or even eliminate features of infrared small-dim targets \cite{liangkui2018using}.

\subsection{Combining self-attention mechanisms with CNNs} 
Inspired by the success of the self-attention mechanism adopted in transformer architectures in the NLP field \cite{vaswani2017attention, ye2021contrastive}, some works employed them in the computer vision field \cite{luo2021dual, li2021two, zhang2020learning}.

Vision Transformer (ViT) is the pioneering work that directly applied a transformer architecture on non-overlapping medium-sized image patches for image classification \cite{dosovitskiy2020image}.
Liu et al. \cite{liu2021swin} proposed a hierarchical Transformer structure named swin transformer.
Such hierarchical architecture had the flexibility to model at various scales and had linear computational complexity with respect to image size.
The most obvious advantage of the transformer is its ability to capture the image information through the self-attention mechanism in a large range.
However, the self-attention mechanism's performance in local information learning is relatively weak compared with CNN-based methods.
Hence, some methods proposed to combine the strengths of CNNs and self-attention mechanisms \cite{liang2020polytransform, dai2021up, zhu2020deformable}. 
Carion et al. \cite{carion2020end} adopted Resnet-50 or Resnet-101 \cite{he2016deep} to acquire the compact feature representation, and then introduced this representation into the self-attention mechanism.
Liu et al. \cite{liu2021swin} proposed a general-purpose transformer backbone for computer vision.
D'Ascoli et al. \cite{DBLP:journals/corr/abs-2103-10697} proposed the gated positional self-attention to mimic the locality of convolutional layers.
Chen et al. \cite{chen2021transunet} combined self-attention mechanism with U-net \cite{ronneberger2015u} for medical segmentation.

Different from these methods, the proposed method in this paper focuses on the small size of the target, and designs a feature enhancement module to learn more discriminative features of small-dim targets. %

\section{Proposed Method} 
\label{sec:method}
\subsection{Overview}
As depicted in Fig. \ref{fig:pipline}, the proposed method consists of three main modules:
(1) A feature embedding module to extract a compact feature representation of an image.  
(2) A compound encoder to learn interaction information amongst all embedded tokens and more discriminative features of small-dim targets. 
(3) A decoder to produce confidence maps.

Given an image of size $C \times H \times W$, we embed it into a sequence of tokens by the Resnet-50 \cite{he2016deep}.
Then the designed compound encoder is used to learn the interaction information amongst all embedded tokens and capture more discriminative features of small-dim targets.
After that, with the help of the U-Net-like \cite{ronneberger2015u} skip connection operation, embedded features are concatenated with feature maps obtained by the decoder to obtain the confidence map.
Finally, we adopt the adaptive threshold \cite{gao2013infrared} to segment the confidence map to obtain the detection result.

\subsection{CNN-based feature embedding module}
In ViT, an image is divided into a sequence of non-overlapping patches of the same size and then use the linear mapping to embed these patches to a sequence of tokens \cite{dosovitskiy2020image}.
In this paper, the proposed method adopts the Resnet-50 \cite{he2016deep} as the feature embedding module to extract compact features of the original image, and then reshapes them into a sequence of tokens.

After the input image $x \in \mathbb{R}^{C \times H \times W}$ passes through the feature embedding module, compact features $F \in \mathbb{R}^{C_1 \times H' \times W'}$ with local information are obtained.
Then we flatten 3D features $F$ into 2D tokens $E_{em} \in \mathbb{R}^{H'W' \times C_1}$, where $H'W'$ is the number of tokens.
To maintain the spatial information of these features, we learn specific position embeddings $E_{pos}$.
Finally, we obtain embedded tokens $E=E_{em}+E_{pos}$, where $E \in \mathbb{R}^{n \times C_1}$ and $E=\left(E_{1}, E_{2}, \cdots, E_{n}\right)$, $n$ is the number of tokens, and $n= {H'W'}$.

\subsection{Compound encoder}
The compound encoder contains $k$ encoder layers, and all encoder layers have the same structure.
An encoder layer includes a multi-head self-attention module with $m$ heads and a feature enhancement module.
The multi-head self-attention mechanism aims to capture the interaction information amongst $n$ tokens so that differences between targets and backgrounds can be well constructed.
The feature enhancement module aims to learn more discriminative features of small-dim targets.

\subsubsection{Multi-head self-attention module}
Embedded tokens $E$ are divided into $m$ heads $E=\left\{E^{1}, E^{2}, \cdots, E^{m}\right\}$, $E^{j} \in \mathbb{R}^{n \times \frac{C_1}{m}}$, and then fed into the multi-head self-attention module $\operatorname{MSA}(\cdot)$ to obtain interaction tokens $E_{a}$, we define these processes as:
\begin{equation}
E_{a}=\operatorname{MSA}\left(\operatorname{LN}\left(E\right)\right)+E,
\end{equation}
where the $\operatorname{LN}(\cdot)$ is the layer normalization.

In each head, the multi-head self-attention module $\operatorname{MSA}(\cdot)$ defines three learnable weight matrices to transform Queries ($W^{Q} \in \mathbb{R}^{ n \times \frac{C_1}{m}}$), Keys ($W^{K} \in \mathbb{R}^{n \times \frac{C_1}{m}}$) and Values ($W^{V} \in \mathbb{R}^{ n \times \frac{C_1}{m}}$).
The embedded tokens $E^j$ of a head are first projected onto these weight matrices to get $Q^j={E^j}{W^Q}$, $K^j={E^j}{W^K}$ and ${V^j}={E^j}{W^V}$. 
The output $Z^j \in \mathbb{R}^{n \times \frac{C_1}{m}}$ of the self-attention layer is given by: 

\begin{equation}
Z^{j}=\operatorname{softmax}\left(\frac{{Q^j}{K^j}^{T}}{\sqrt{\frac{C_1}{m}}}\right){V^j},
\label{msa}
\end{equation}
where $j$ is the $j$-th head of the multi-head self-attention. 
The result of $m$ heads can be expressed as: 

\begin{equation}
Z=\left\{Z^{1}, Z^{2}, \cdots, Z^{m}\right\}, Z \in \mathbb{R}^{n \times C_1}
\end{equation}

\subsubsection{Feature enhancement module}
We feed interaction tokens $E_{a}$ into the designed feature enhancement module to obtain compound tokens $E_{a\_l}$.

Specifically, the feature enhancement module is shown in Fig. \ref{fig:pipline}.
Firstly, these interaction tokens $E_{a}$ are fed into the first full connection layer to obtain 2D tokens $F_a=\left(F_{a\_1}, F_{a\_2}, \cdots, F_{a\_n}\right)$,
$F_{a} \in \mathbb{R}^{n \times C_2}$. 
Then we reshape 2D tokens into 3D features $F_{3D}$ with the size of $n \times P \times P$ and adopt the convolution operation to learn the local information of $F_{3D}$, which helps enhance the features of small-dim targets.
Finally, enhanced features $F_{conv}$ are further reshaped back to the size of ${n \times C_2}$ and then fed into the next full connection layer to learn the next 2D tokens $F_{a\_l}$.
After that, with the summation of $F_{a\_l}$ and $E_{a}$, we obtain compound tokens $E_{a\_l} \in \mathbb{R}^{n \times C_1}$.

\subsection{Feature decoder with skip connection} %
To obtain confidence maps of small-dim targets, we adopt a decoder to upsample reshaped compound tokens $F_{up}\in \mathbb{R}^{ H' \times W' \times C_1}$.
To prevent the loss of the small-dim target contextual information, all features in the feature embedding process are concatenated with feature maps obtained by upsampling operation through skip connection operation like U-Net structure \cite{ronneberger2015u}.
Then, the confidence map $y$ is obtained by the sigmoid function.
Finally, we adopt the adaptive threshold \cite{gao2013infrared} for target segmentation, and achieve the detection result. 

\subsection{Loss function}
To handle the class imbalance issue between targets and backgrounds \cite{dai2021attentional} and focus more on small-dim target regions, the Intersection of Union (IoU) loss is adopted to calculate the distance between the confidence map and the ground truth, defined by:

\begin{equation}
\label{iouloss}
L_{i o u}=1-\frac{y \cap x_{gt}}{y \cup x_{gt}}.
\end{equation}

The $y$ is the confidence map, and the $x_{gt}$ is the ground truth image.

\section{Experiment} 
\label{sec:experiment}
In this section, we first introduce datasets, evaluation metrics, and implementation details, respectively.
Then we compare the proposed method with state-of-the-art methods.
Finally, we conduct the ablation study, evaluate the performance on cross-scene generalization, and verify the proposed method has a stronger anti-noise performance.

\subsection{Experimental setup}
\subsubsection{Datasets} 
We adopt the widely used MFIRST dataset \cite{wang2019miss} and SIRST dataset \cite{dai21acm} to evaluate the proposed method. 
The MFIRST dataset contains 9960 training samples and 100 test samples.
Among them, all infrared small-dim target image samples are generated by the random combination of real backgrounds and real small-dim targets or simulated targets with Gaussian spatial gray distribution.
The SIRST dataset is a widely-used public dataset that contains 341 training samples and 86 test samples.

\begin{figure}[]
    \centering
    \includegraphics[width =0.5\textwidth]{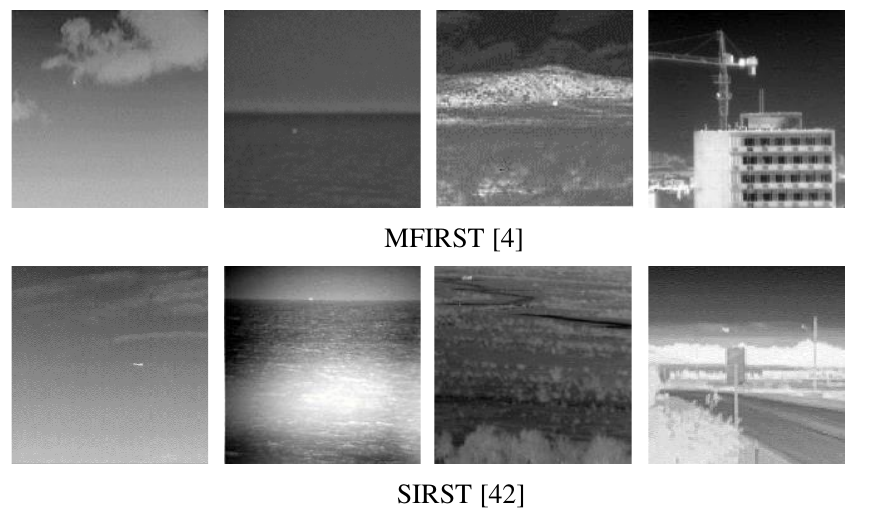}
    \caption{Representative samples of datasets.}
    \label{fig:dataset}
 \end{figure}

Fig.\ref{fig:dataset} shows representative images of these two datasets.
In these datasets, small-dim targets usually appear in the sea, sky, mountains, or buildings background.
Compared with MFIRST dataset, small-dim targets in SIRST dataset have some shape information.

\begin{table*}[h]
\centering
\caption{Comparison on different datasets. `-' means that the method can not get reasonable values under fixed $F_a=0.2$ for $P_d$ or under $F_a\leqslant2.0$ for AUC.}
\resizebox{\textwidth}{!}{%
\begin{tabular}{c|cccc|cccc|c}
\hline
\multirow{2}{*}{Methods} & \multicolumn{4}{c|}{MFIRST} & \multicolumn{4}{c|}{SIRST} & \multirow{2}{*}{\begin{tabular}[c]{@{}c@{}}Times\\ (s/100 images)\end{tabular}} \\ \cline{2-9}
                         & $P_d$(\%)                    & AUC(\%)        & $F_{1}^t$(\%)      & \multicolumn{1}{c|}{$F_{1}^p$(\%)}                                & $P_d$(\%)               & AUC(\%)              & $F_{1}^t$(\%)            & $F_{1}^p$(\%)  &                                                                                 \\ \hline 
\multicolumn{1}{c|}{Top-Hat \cite{zeng2006design}}                              & -                         & -              & 44.62          & \multicolumn{1}{c|}{12.8}                                     & 85.34                & 82.38                & 82.52                & 44.13              & 1.78       \\
\multicolumn{1}{c|}{Max-Mean/Max-Media \cite{deshpande1999max}}                             & -                         & 50.50          & 58.30          & \multicolumn{1}{c|}{14.44}                                    & 78.46                & 77.45                & 73.49                & \multicolumn{1}{c|}{23.97}              &1.50        \\
\multicolumn{1}{c|}{AAGD \cite{aghaziyarati2019small}}                                 & 43.66                     & 56.96          & 65.70          & \multicolumn{1}{c|}{32.42}                                    & 89.09                & 88.14                & 84.69                & \multicolumn{1}{c|}{50.27}               &3.52       \\
\multicolumn{1}{c|}{ADMD \cite{moradi2020fast}}                                 & 59.64                     & 64.09          & 70.99          & \multicolumn{1}{c|}{31.52}                                    & 94.13                & 90.46                & 88.50                & \multicolumn{1}{c|}{56.69}                  &2.02    \\
\multicolumn{1}{c|}{LIG \cite{zhang2018infrared}}                                  & 59.29                     & 64.17          & 70.87          & \multicolumn{1}{c|}{41.27}                                    & 90.19                & 90.00                & 89.72                & \multicolumn{1}{c|}{59.15}              & 70.44       \\
\multicolumn{1}{c|}{IPI \cite{gao2013infrared}}                                  & 41.59                     & 51.02          & 60.73          & \multicolumn{1}{c|}{33.58}                                    & 86.87                & 84.45                & 85.32                &\multicolumn{1}{c|}{56.97}               &424.60        \\
\multicolumn{1}{c|}{ILCM \cite{han2014robust}}                                 & -                         & -              & 24.52          & \multicolumn{1}{c|}{0.91}                                     & -                    & -                    & 47.26                & \multicolumn{1}{c|}{0.71}              &1.92         \\
\multicolumn{1}{c|}{MPCM \cite{wei2016multiscale}}                                 & 57.86                     & 64.62          & 72.20          & \multicolumn{1}{c|}{35.43}                                    & 93.56                & 90.40                & 86.96                & \multicolumn{1}{c|}{58.59}             &4.60         \\
\multicolumn{1}{c|}{TLLCM \cite{han2019local}}                                & -                         & 46.43          & 52.63          & \multicolumn{1}{c|}{6.67}                                     & 61.61                & 79.14                & 79.66                & \multicolumn{1}{c|}{7.60}              &321.91         \\
\multicolumn{1}{c|}{LEF \cite{xia2019infrared}}                                  & 49.49                     & 70.01          & 72.45          & \multicolumn{1}{c|}{5.87}                                     & -                    & -                    & 59.60                & \multicolumn{1}{c|}{2.45}              &430.22         \\
\multicolumn{1}{c|}{GST \cite{gao2008generalised}}                                  & 56.39                     & 59.69          & 66.67          & \multicolumn{1}{c|}{24.67}                                    & 77.01                & 76.81                & 80.40                & \multicolumn{1}{c|}{35.32 }            &1.05         \\ \hline
\multicolumn{1}{c|}{MDvsFA \cite{wang2019miss}}                               & 86.62                     & 81.78          & 85.27          & \multicolumn{1}{c|}{60.36}                                    & -                & -                & -                & \multicolumn{1}{c|}{-}             &10.62         \\
\multicolumn{1}{c|}{ACM \cite{dai21acm}}   & 70.07                     & 71.95          & 82.11          & \multicolumn{1}{c|}{58.05}   & 98.24                & 91.67                & 96.78                & \multicolumn{1}{c|}{81.30}           &1.61           \\ \hline
\multicolumn{1}{c|}{Ours}  & \textbf{90.08}            & \textbf{89.34} & \textbf{92.59} & \multicolumn{1}{c|}{\textbf{64.59}}                           & \textbf{100.00}        & \textbf{99.14}       & \textbf{98.62}       & \multicolumn{1}{c|}{\textbf{83.16}}      &6.41       \\ \hline

\end{tabular}%
}
\label{tab:comparison on different datasets}
\end{table*}

\begin{figure*}[h]
    \centering
    \includegraphics[width =1\textwidth]{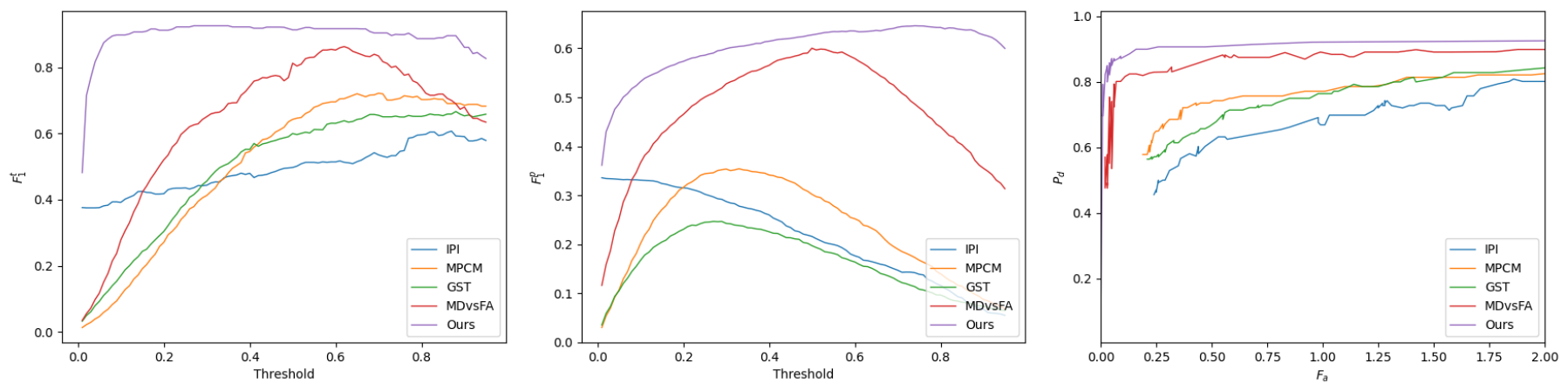}
    \caption{The $F_{1}^t$ and $F_{1}^p$ curves with different thresholds and the $P_d-F_a$ ROC curve on MFIRST dataset.
    }
    \label{fig:curve}
 \end{figure*}

\subsubsection{Evaluation metrics}
As the same as \cite{gao2013infrared}, we regard that the detection is correct when the following two conditions are met simultaneously: 
(1) The output result has some overlap pixels with the ground truth. 
(2) The pixel distance between the centers of the detection result and the ground truth is less than a threshold (4 pixels).

In this paper, four widely used evaluation metrics \cite{gao2013infrared, wang2019miss}, including the Probability of detection ($P_d$), False alarm ($F_a$) rate, target-level $F_1$ measure ($F_{1}^t$) and pixel-level $F_1$ measure ($F_{1}^p$) are used for performance evaluation.
These evaluation metrics are defined as follows:

\begin{equation}
P_{d}=\frac{ \# \text { number of true detections }}{\# \text { number of real targets }},
\end{equation}

\begin{equation}
F_{a}=\frac{ \# \text {number of false detections }}{\# \text { number of images }}.
\end{equation}

As the same as \cite{gao2013infrared}, we adopt $P_d$ with $F_a$=0.2 and the area under $P_d-F_a$ curve (AUC) with $F_a < 2.0$ to evaluate average performance of the proposed method.

\begin{equation}
F_{1}^t=\frac{2\times Precision_t\times Recall_t}{Precision_t + Recall_t},
\end{equation}
\begin{equation}
F_{1}^p=\frac{2\times Precision_p\times Recall_p}{Precision_p + Recall_p},
\end{equation}
where the precision and recall can be defined as follows:
\begin{equation}
Precision=\frac{TP }{TP+FP},
\end{equation}
\begin{equation}
Recall=\frac{TP }{TP+FN},
\end{equation}
where $TP$ is the true positive, $FP$ is the false positive and $FN$ is the false negative.

The $P_d$ can measure the correct detection rate of the model when $F_a=0.2$.
The AUC can reflect the false alarm rate of the model.
A model with a low false alarm rate will have a higher AUC value.

\subsubsection{Implementation details}  
The framework of the proposed method is implemented using Pytorch 1.7.1, and accelerated by CUDA 11.2.
The whole network is trained with the SGD algorithm with a learning rate of 0.01, momentum 0.9, and weight decay 1e-4 on NVIDIA GeForce RTX 3090 GPU.
All trainable methods are trained from the scratch and the batch size is set to 24. 
The input image is resized to $224 \times 224$. 
In terms of the results of the hyperparameter discussion in the ablation study, the head of the MSA is set to 12, and the number of the encoder layer is set to 12. 

\subsection{Comparison with the state-of-the-art methods}
We compare the proposed method with following related methods:
\begin{itemize}
   \item Model-driven methods: Top-Hat \cite{zeng2006design}, Max-Mean/Max-Median \cite{deshpande1999max}, AAGD \cite{aghaziyarati2019small}, ADMD \cite{moradi2020fast}, LIG \cite{zhang2018infrared}, IPI \cite{gao2013infrared}, ILCM \cite{han2014robust}, MPCM \cite{wei2016multiscale}, TLLCM \cite{han2019local}, LEF \cite{xia2019infrared}, GST \cite{gao2008generalised}.
    \item Deep learning methods: MDvsFA \cite{wang2019miss} and ACM \cite{dai21acm}.
\end{itemize}

Since the open-source code of the deep-learning-based infrared small-dim target detection method is limited, we adopt the open-source MDvsFA and ACM methods for performance comparison.

\begin{figure*}[h]
\centerline{\includegraphics[width =1\textwidth]{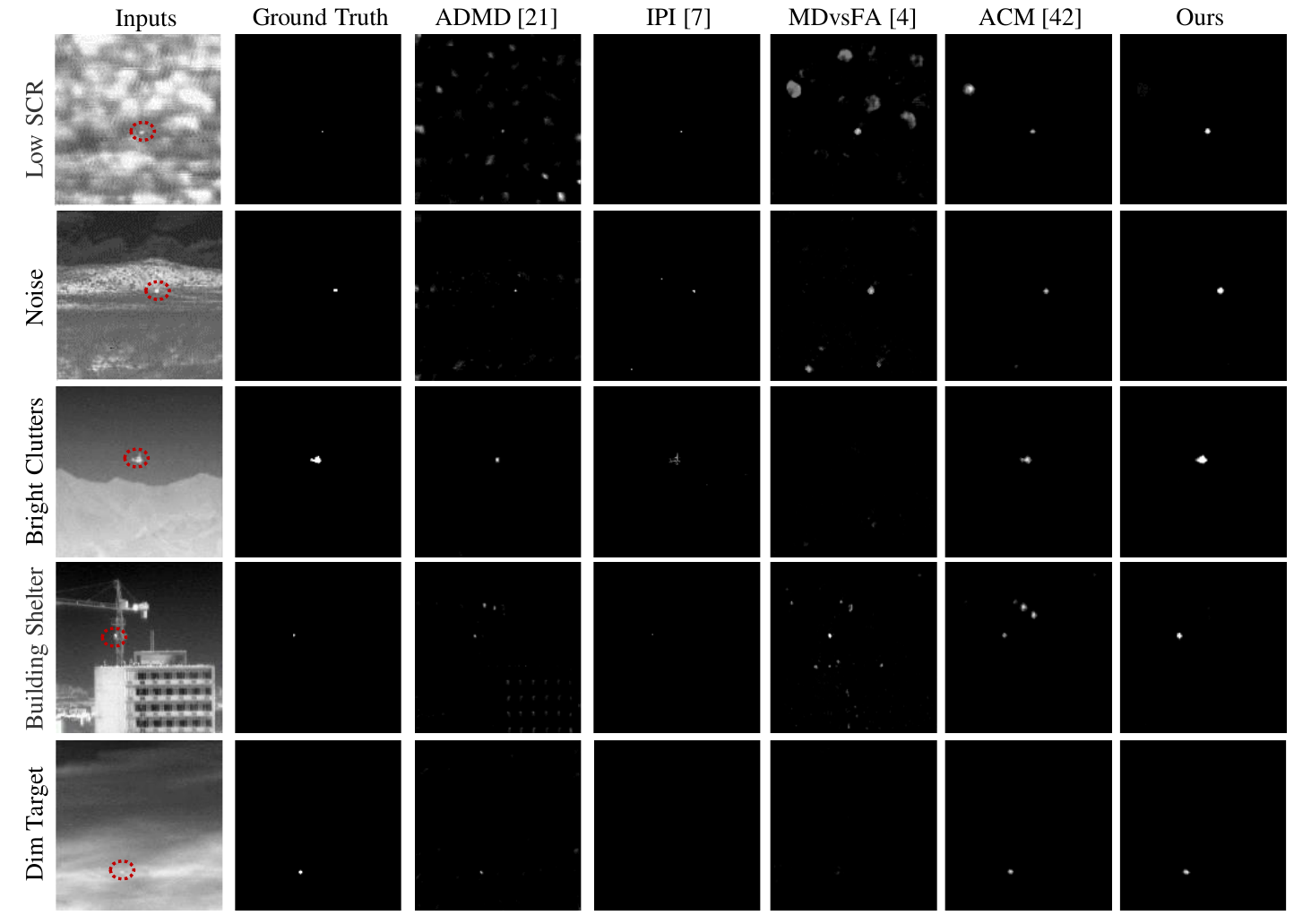}}
\caption{
 The representative processed results of different methods.
 The first row to the last row shows the results of detecting a small-dim target with a low SCR, with a high noise level, with bright clutters, with complex building shelters, and with a dim appearance, respectively.
}
\label{fig:visualization of likelihood maps from six methods}
\end{figure*}

\subsubsection{Quantitative evaluation}
The results of different methods on two public datasets \cite{wang2019miss, dai21acm} are listed in Table \ref{tab:comparison on different datasets}.
From this table, we can observe that deep-learning-based methods are significantly superior to model-driven methods.
The proposed method outperforms both model-driven methods and deep-learning-based methods.

On the MFIRST dataset, compared with the MDvsFA method, the proposed method outperforms approximately 3.46\% on $P_d$, 7.56\% on AUC, 7.32\% on $F_{1}^t$ and 4.23\% on $F_{1}^p$. 
Because the MDvsFA method is difficult to balance outputs of two detection generators (There are two generators $G_1$ and $G_2$. $G_1$ aims to minimize MD, $G_2$ aims to minimize FA), and this method does not focus on exploring the difference between the target and the background.
Compared with the ACM method, the proposed method also outperforms a lot.
The ACM approach can not adequately learn the interaction information amongst image features, because the locality of CNNs impairs the ability to capture large-range dependencies. 
The proposed method not only learns the interaction information amongst all embedded tokens, but also extracts more discriminative features of small-dim targets from a local view.
Consequently, the proposed method obtains the best performance.

On the SIRST dataset, compared with the state-of-the-art ACM method, the proposed method outperforms approximately 1.76\% on $P_d$, 7.47\% on AUC, 1.84\% on $F_{1}^t$ and 1.86\% on $F_{1}^p$.
The ACM method is limited by the locality of convolutional neural networks, while the proposed method adopts the self-attention mechanism which can learn the relationships of one token with others in a larger region so that the interaction information amongst all embedded features of an image can be better constructed.
It is worth noting that the GANs-based MDvsFA method is difficult to converge on SIRST datasets, so we do not show the detection result.

The $F_{1}^t$, $F_{1}^p$ and $P_d-F_a$ ROC curves of some representative methods on MFIRST dataset are shown in Fig. \ref{fig:curve}.
As we can see from the figure, the proposed method always has better performance than other methods at different thresholds.
These results validate the robustness of the proposed method.

\subsubsection{Qualitative evaluation}
Fig. \ref{fig:visualization of likelihood maps from six methods} shows qualitative evaluation comparisons of five representative methods.
It can be seen that all targets in infrared images with different complex backgrounds appear small-dim and sparse, but the proposed method can achieve the best detection performance.
When detecting the small-dim target with a high noise level, a low SCR, and with building shelters, most methods, including ADMA, MDvsFA, and ACM have some false alarms.
In these cases, the local regions similar to infrared small-dim targets spread over the whole background, and these methods do not focus on exploring the difference between the small-dim target and the background.
When detecting the small-dim target with bright clutters, the MDvsFA method fails to detect the right target.
When the infrared target is pretty dim and the contrast between the background and the target is extremely weak, some methods are difficult to acquire robust detection.
This can be seen in the last row of Fig. \ref{fig:visualization of likelihood maps from six methods} where the small-dim target is hidden in the clouds.
These methods can not learn the difference between the small-dim target and the background adequately. 
Besides, they may lose some discriminative feature of small-dim targets during feature extraction.
In contrast, the proposed method can learn the interaction information amongst embedded tokens in a larger range, and focus on learning more discriminative features of small-dim targets, so it has the best detection results in all kinds of complex backgrounds.

\begin{table}[t]
\centering
\caption{The performance of the module ablation on different datasets.}
\resizebox{1\columnwidth}{!}{
\begin{tabular}{c|cc|cc}
\hline
\multirow{2}{*}{Methods} & \multicolumn{2}{c|}{MFIRST}                                       & \multicolumn{2}{c}{SIRST}                                                    \\ \cline{2-5} 
                         & $P_d$(\%)         & AUC(\%)         & $P_d$(\%)        & AUC(\%)                         \\ \hline
Baseline                 & 80.00          & 80.76                  & 99.78         & 97.98                      \\
Baseline-pool            & 87.05          & 84.77                  & 99.90         & 99.00                       \\
\begin{tabular}[c]{@{}c@{}}Baseline-pool+FEM \\ (Ours)\end{tabular}        & \textbf{90.08}          & \textbf{89.34}              & \textbf{100.00}          & \textbf{99.14}  \\ \hline
\end{tabular}%
}
\label{tab:ablation}
\end{table}

\subsection{Ablation study}
In this section, we validate the effect of the compound encoder which helps learn the interaction information amongst all embedded tokens and more discriminative features of small-dim targets.
The results are shown in Table \ref{tab:ablation}.
The `baseline' method \cite{chen2021transunet} for comparison consists of three modules: a Resnet-50 structure with a pooling layer \cite{he2016deep}, a self-attention encoder that has the same structure as the ViT \cite{dosovitskiy2020image} model, a decoder with skip connection.
The `baseline-pool' method removes the pooling layer from the Resnet-50 structure, and only retains the IoU loss.
The `baseline-pool+FEM' method is the proposed method in this paper.

Experimental results in Table \ref{tab:ablation} obviously show the promotion effect of each component on infrared small-dim target detection.
As can be seen, the baseline method with the normal transformer encoder can achieve reasonable results.
When we remove the pooling layer, all metrics improve a lot.
These results show that the pooling layer can degrade the feature learning of small-dim targets, which is the same as \cite{liangkui2018using}.
When we adopt a feature enhancement module (FEM) in the compound encoder with the self-attention mechanism, the performances are further improved.
On MFIRST dataset, the $P_d$ is improved by 3.03\%, and the AUC is improved by 4.57\%. 
On SIRST dataset, the $P_d$ is improved by 0.10\%, and the AUC is improved by 0.14\%.
It shows that the self-attention mechanism helps capture the interaction information amongst embedded tokens to learn the difference between the target and background from a large region.
Furthermore, the local information learned by the feature enhancement module helps learn more discriminative features of small-dim targets.

\subsection{Hyperparameter discussion}
The number of encoder layers and the head number of the MSA module influence the effect of learning the interaction information amongst all embedded tokens. 
In this section, we investigate these important hyperparameters, the results can be seen in Table \ref{tab:Hyperparameter ablation}.

As can be observed in Table \ref{tab:Hyperparameter ablation}, as the number of encoder layers increases, the time consumption of detection increases.
As the number of encoder layers decreases, the dependency between the target and background can not be constructed adequately.
The head number of MSA has little effect on performance.
Finally, based on the best experimental performance, we set the number of the encoder layers to 12, and the head number of the MSA module to 12.

\begin{table}
\centering
\caption{The performance of the hyperparameter ablation on the MFIRST dataset}
\setlength{\tabcolsep}{4mm}{
\begin{tabular}{cc|ccc}
\hline
\begin{tabular}[c]{@{}c@{}}Encoder \\ layers\end{tabular} & \multicolumn{1}{c|}{Head} & $P_d$(\%)         & AUC(\%)        & \begin{tabular}[c]{@{}c@{}}Times\\ (s/100 images)\end{tabular} \\ \cline{1-5}
14                                            & \multicolumn{1}{c|}{3}    & 84.39          & 84.38          & 6.62                                                          \\
14                                            & \multicolumn{1}{c|}{12}   & 85.71          & 85.08          & 6.61                                                            \\
12                                            & \multicolumn{1}{c|}{3}    & 87.14          & 83.11          & 6.41                                                   \\
12                                            & \multicolumn{1}{c|}{12}   & \textbf{90.08} & \textbf{89.34} & 6.41                                                            \\
6                                             & \multicolumn{1}{c|}{3}    & 85.00          & 84.26          & 5.43                                                            \\
6                                             & \multicolumn{1}{c|}{12}   & 85.71          & 85.52          & 5.33                                                            \\ \cline{1-5}
\end{tabular}%

\label{tab:Hyperparameter ablation}
}
\end{table}

\subsection{Cross-scene generalization}
In practice, the well-trained model is likely to be applied to a new scene, so the cross-scene generalization is very important.
Consequently, we evaluate the generalization of the proposed model on cross-scene datasets.
The experimental results are listed in Table \ref{tab:Generalization ability}.

As can be seen from Table \ref{tab:Generalization ability}, these experiments verify the strong generalization of deep learning methods.
However, compared with other deep learning methods, the proposed method achieves better performance in terms of cross-scene generalization.
When the model is trained on the MFIRST dataset and tested on the SIRST dataset, the proposed method outperforms approximately 0.93\% on $P_d$ and 2.07\% on AUC, compared with the state-of-the-art MDvsFA method.
When the model is trained on the SIRST dataset and tested on the MFIRST dataset, the proposed method outperforms approximately 20.52\% on $P_d$ and 23.53\% on AUC, compared with the state-of-the-art ACM method.
Although the distribution of the dataset is heterogeneous, the differences between infrared small-dim targets and background in different datasets are similar.
The self-attention mechanism helps capture the interaction information amongst embedded tokens to learn the difference between the target and background from a large region, so it has the best detection results under the cross-scene situation.

\begin{table}
\centering
\caption{The performance of different methods in terms of cross-scene generalization. `-' means that the method can not get reasonable values.}
\setlength{\tabcolsep}{4.1mm}{
\begin{tabular}{c|cc|cc}
\hline
\multirow{2}{*}{Methods} & \multicolumn{2}{c|}{\begin{tabular}[c]{@{}c@{}}MFIRST(Train)\\ SIRST(Test)\end{tabular}} & \multicolumn{2}{c}{\begin{tabular}[c]{@{}c@{}}SIRST(Train)\\ MFIRST(Test)\end{tabular}} \\ \cline{2-5} 
                         & $P_d$(\%)               & AUC(\%)             & $P_d$(\%)               & AUC(\%)             \\ \hline
MDvsFA \cite{wang2019miss}                  & 97.22                & 93.95               & -                 & -               \\
ACM \cite{dai21acm}                     & 92.13                & 57.49               & 67.25                    & 62.05               \\
ours      & \textbf{98.15}       & \textbf{96.02}      & \textbf{87.77}       & \textbf{85.58}      \\ \hline
\end{tabular}%

\label{tab:Generalization ability}
}
\end{table}

\begin{figure}[b]
\centerline{\includegraphics[width =0.50\textwidth]{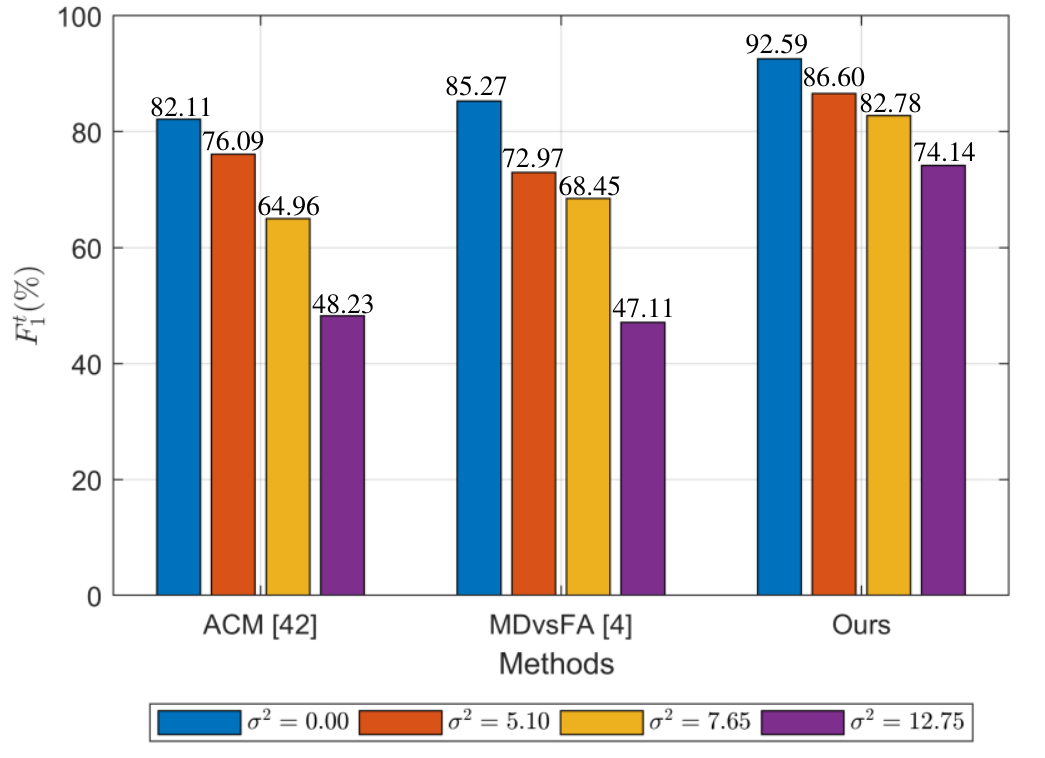}}
\caption{
 The $F_1^t$ performance of different methods under Gaussian noises with different variances.
`$\sigma^2$' means the variance of Gaussian noise.}
\label{fig:f1t}
\end{figure}

\begin{figure}[]
\centerline{\includegraphics[width =0.50\textwidth]{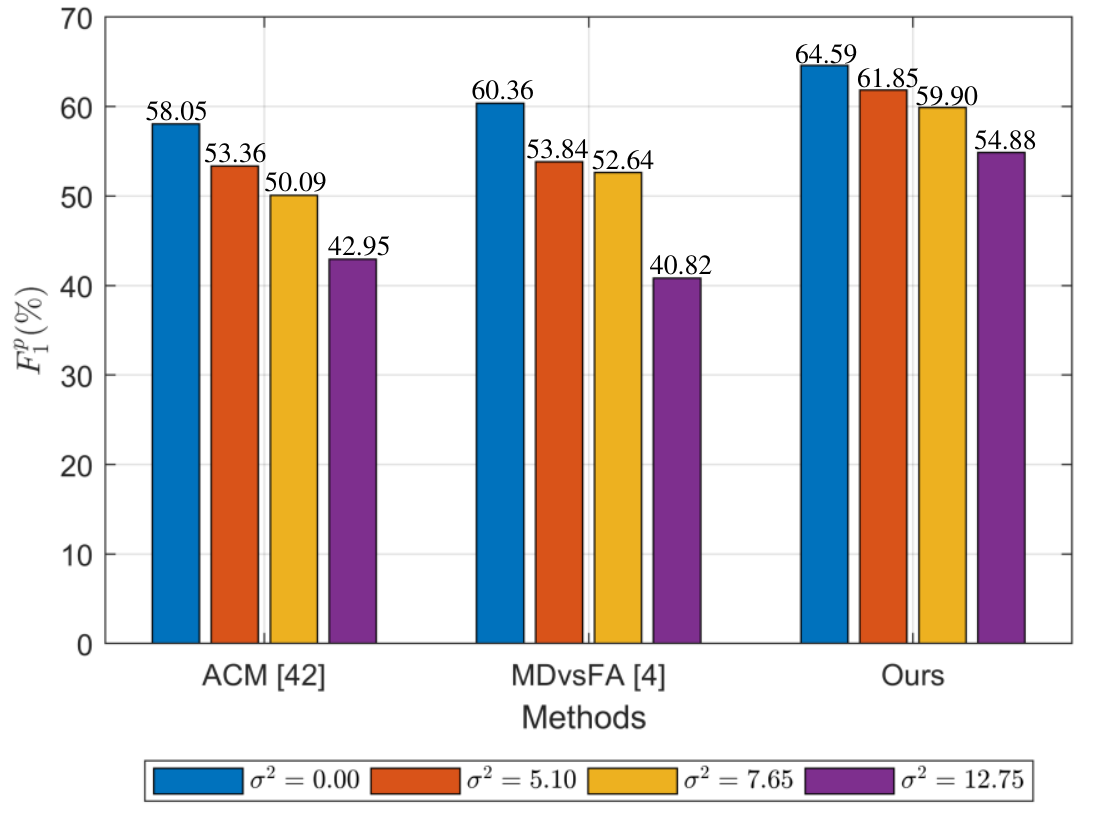}}
\caption{
 The $F_1^p$ performance of different methods under Gaussian noises with different variances.
}
\label{fig:f1p}
\end{figure}

\begin{figure}[]
\centerline{\includegraphics[width =0.50\textwidth]{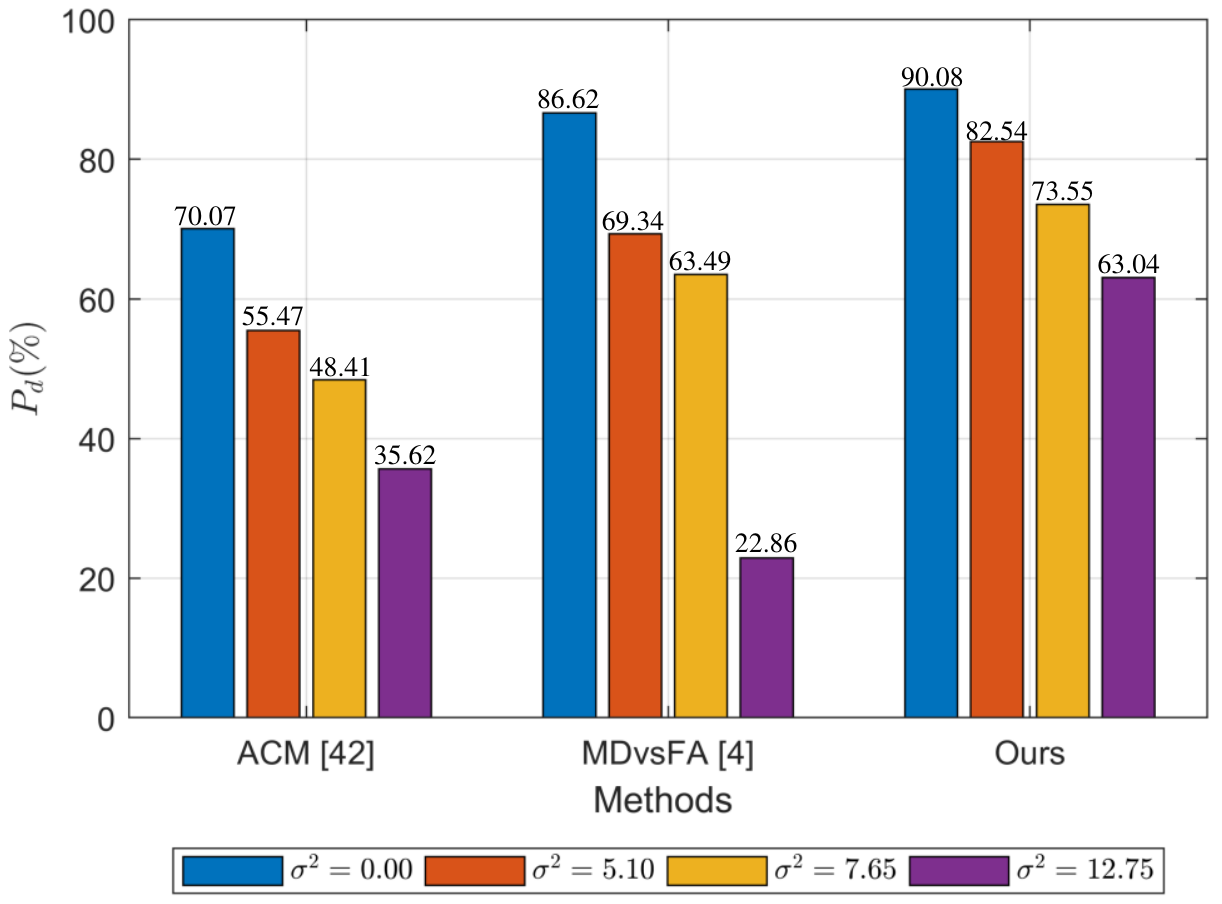}}
\caption{
 The $P_d$ performance of different methods under Gaussian noises with different variances.
}
\label{fig:pd}
\end{figure}

\begin{figure}[]
\centerline{\includegraphics[width =0.50\textwidth]{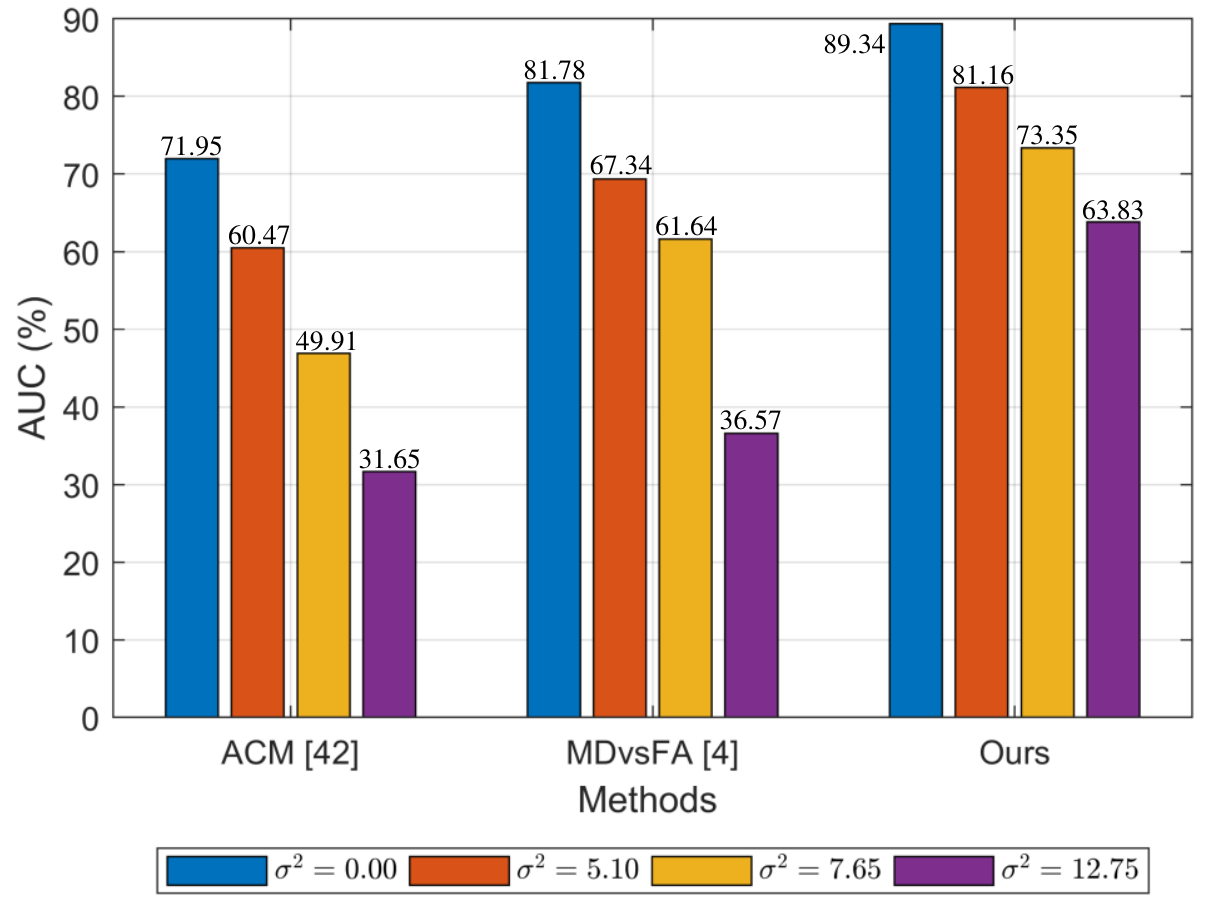}}
\caption{
 The AUC performance of different methods under Gaussian noises with different variances.
}
\label{fig:auc}
\end{figure}

\subsection{Evaluation of anti-noise performance}
In addition to cross-scene generalization, the anti-noise performance is also very crucial.
Thus, we evaluate the anti-noise performance of the proposed model on the MFIRST dataset.
We set four kinds of Gaussian noise with different variances, which are 5.10, 7.65, 12.75, and 25.50, respectively.
The mean of these Gaussian noises is set to 0.
It's worth noting that we only add the above noise to the test set.

When the variances are 0.00, 5.10, 7.65, and 12.75, the experimental results are showed in Fig. \ref{fig:f1t}, \ref{fig:f1p}, \ref{fig:pd}, and \ref{fig:auc}. 
As we can see from these figures, the detection performance decreases with the increase of noise.
However, compared with other deep learning methods, the proposed method has the best performance under Gaussian noise with different variances.
Meanwhile, with the increase of noise variance, the performance of the proposed method decreases relatively slowly over other methods.
These experiments verify the proposed method has stronger anti-noise performance.

\begin{figure}[]
\centerline{\includegraphics[width =0.50\textwidth]{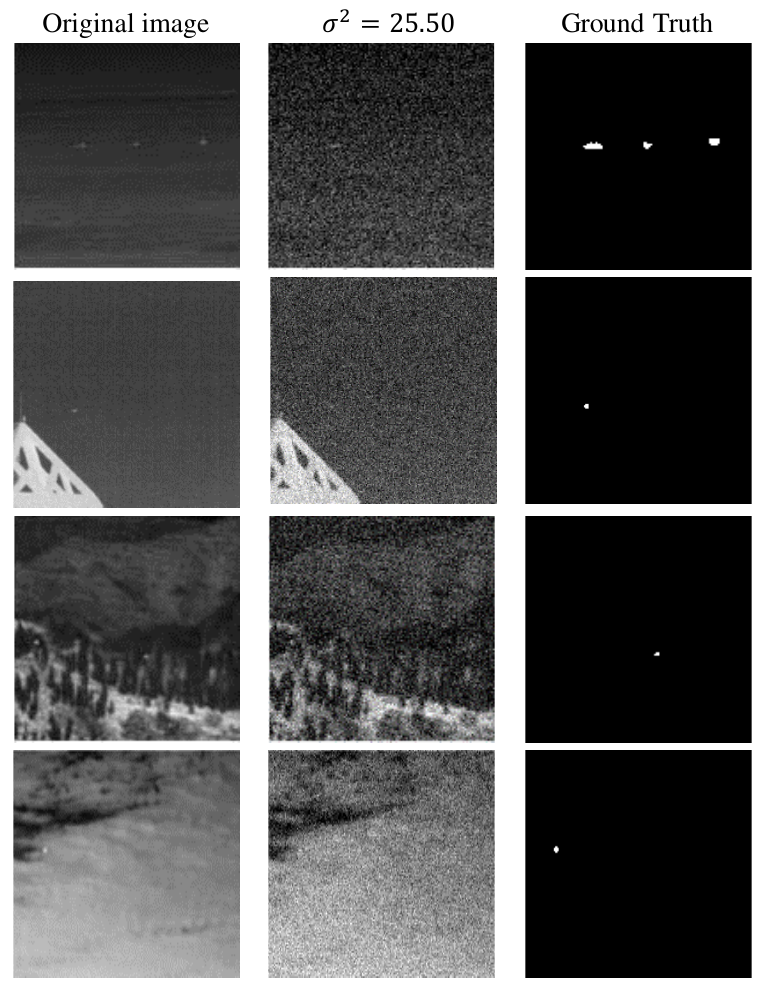}}
\caption{From left to right, the first column is the original image, the second column is the image after adding Gaussian noise, and the third column is the ground truth.
`$\sigma^2$' means the variance of Gaussian noise.}
\label{fig:noise}
\end{figure}

\begin{figure}[]
\centerline{\includegraphics[width =0.50\textwidth]{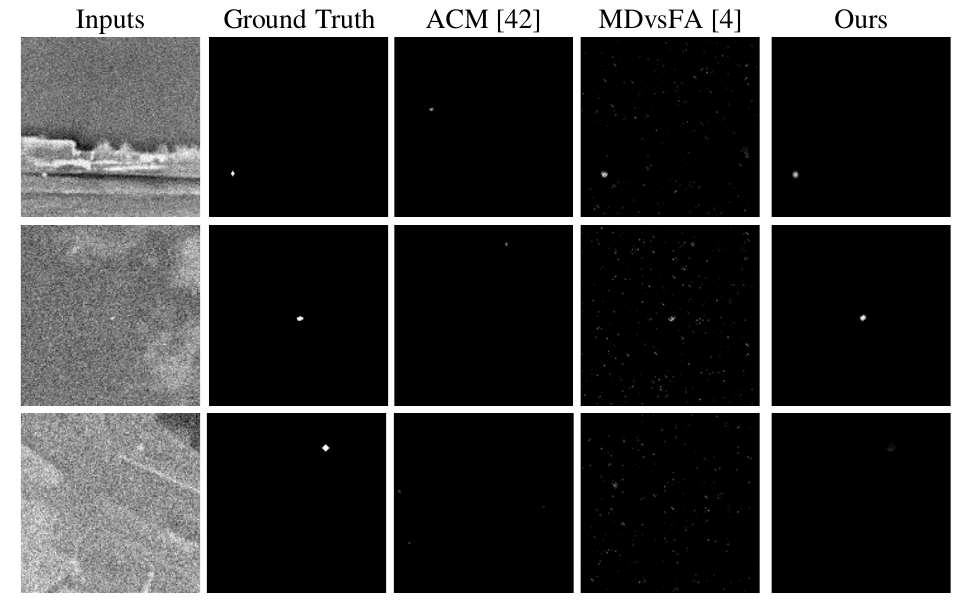}}
\caption{The detection performance when the variance of the Gaussian noise is 25.50.}
\label{fig:010compare}
\end{figure}

When the variances is 25.50, the Fig. \ref{fig:noise} illustrates some samples of inputs.
As we can see from this figure, the small-dim targets are completely submerged in backgrounds, and we can hardly find out small-dim targets.
The Table \ref{tab:noise010} shows the detection performance of different methods, and we can see from the table that the performance of all methods deteriorates dramatically.
Even so, the proposed method still has the best experimental performance.
Fig. \ref{fig:010compare} shows the detection performance of different methods when the variance of Gaussian noise is 25.50.
As we can observe from the figure, when the small-dim target is not submerged in backgrounds completely, the ACM method tends to miss detection, while the MDvsFA network tends to increase false alarms.
In contrast, the proposed method can robustly detect small-dim targets.

\begin{table}[]
\centering
\caption{The detection performance of different methods when the variance of the Gaussian noise is 25.50.}
\setlength{\tabcolsep}{4.1mm}{
\begin{tabular}{c|cccc}
\hline
Methods & $F_{1}^t$(\%)            & $F_{1}^p$(\%)         & $P_d$             & AUC            \\ \hline
ACM \cite{dai21acm}     & 22.76         & 22.65          & 13.90           & 13.85          \\
MDvsFA \cite{wang2019miss}  & 24.05         & 23.04          & 18.81          & 18.37          \\
Ours     & \textbf{47.40} & \textbf{41.67} & \textbf{34.88} & \textbf{35.67} \\ \hline
\end{tabular}
}
\label{tab:noise010}
\end{table}

\section{Conclusion}
\label{sec:conclude}
In this paper, we propose a new infrared small-dim target detection framework.
We adopt the multi-head self-attention module to explore the interaction information amongst all embedded tokens, and thus differences between targets and backgrounds can be well learned. 
In addition, the feature enhancement module is designed to learn more discriminative features of small-dim targets.
Experiments on two public datasets show that compared with state-of-the-art methods, the proposed method performs much better on detecting infrared small-dim targets with complex backgrounds.
Additionally, experimental results also show the proposed method has a stronger generalization and a better anti-noise performance.

\bibliographystyle{ieeetr}
\bibliography{ref}

\end{document}